\documentclass{article}

\PassOptionsToPackage{numbers, compress}{natbib}

\usepackage{graphicx}

\usepackage[final]{neurips_2024}

\usepackage[utf8]{inputenc} 
\usepackage[T1]{fontenc}    
\usepackage{hyperref}       
\usepackage{url}            
\usepackage{booktabs}       
\usepackage{amsfonts}       
\usepackage{nicefrac}       
\usepackage{microtype}      
\usepackage{xcolor}         

\usepackage{comment}
\usepackage{multirow}
\usepackage{subfig} 
\usepackage{subcaption}
\usepackage{wrapfig,lipsum,booktabs}

\title{Hollowed Net for On-Device Personalization of Text-to-Image Diffusion Models}

\author{Wonguk Cho\thanks{Work done during an internship at Qualcomm AI Research.} $^{\,,1,2}$ \quad Seokeon Choi$^1$ \quad Debasmit Das$^1$ \quad Matthias Reisser$^1$\\ 
\textbf{Taesup Kim}$^2$ \quad \textbf{Sungrack Yun}$^1$ \quad \textbf{Fatih Porikli}$^1$\\\\
$^1$Qualcomm AI Research\thanks{Qualcomm AI Research is an initiative of Qualcomm Technologies, Inc.} \quad $^2$Seoul National University\\\\
$^1${\texttt{\small\{wongcho, seokchoi, debadas, mreisser, sungrack, fporikli\}@qti.qualcomm.com}}\\
$^2${\texttt{\small\{wongukcho, taesup.kim\}@snu.ac.kr}}\\
}

\begin{document}

\maketitle

\begin{abstract}
Recent advancements in text-to-image diffusion models have enabled the personalization of these models to generate custom images from textual prompts. This paper presents an efficient LoRA-based personalization approach for on-device subject-driven generation, where pre-trained diffusion models are fine-tuned with user-specific data on resource-constrained devices. Our method, termed Hollowed Net, enhances memory efficiency during fine-tuning by modifying the architecture of a diffusion U-Net to temporarily remove a fraction of its deep layers, creating a \textit{hollowed} structure. This approach directly addresses on-device memory constraints and substantially reduces GPU memory requirements for training, in contrast to previous methods that primarily focus on minimizing training steps and reducing the number of parameters to update. Additionally, the personalized Hollowed Net can be transferred back into the original U-Net, enabling inference without additional memory overhead. Quantitative and qualitative analyses demonstrate that our approach not only reduces training memory to levels as low as those required for inference but also maintains or improves personalization performance compared to existing methods.

\end{abstract}

\section{Introduction}

Recent research on text-to-image (T2I) diffusion models~\cite{ho2020denoising,rombach2022high}, which generate high-resolution images from text prompts, has increasingly focused on personalizing and customizing these generative models effectively~\cite{gal2022image,zhang2023adding,ruiz2023dreambooth,shah2023ziplora,kumari2023multi}. A primary approach, termed subject-driven generation~\cite{ruiz2023dreambooth}, involves fine-tuning pre-trained diffusion models with a few user-specific images to generate varied representations of a subject using simple text prompts. This allows users to create personalized images of specific subjects, such as family, friends, pets, or personal items, with preferred appearances, backgrounds, and styles. Such capabilities enable creative applications including art renditions, property modifications, and accessorization.

From a practical standpoint, implementing subject-driven generation on-device offers significant benefits in efficiency and privacy. By operating independently of congested cloud servers or networks, users can generate personalized images anywhere at no additional cost and do not need to compromise their privacy as all data and personal information remain on the device.

Despite extensive research aimed at efficiently personalizing diffusion models, limited attention has been paid to memory I/O, a critical bottleneck in on-device learning. Recent studies have mainly explored two strategies: (1) decreasing the number of training steps and (2) reducing the number of updating parameters. The first methods~\cite{ruiz2023hyperdreambooth, gal2023encoder,wei2023elite, chen2024subject,li2024blip} utilize additional large pre-trained models to generate a set of personalized Low-Rank Adaptation (LoRA) parameters~\cite{hu2021lora}, text embeddings, or image prompts from a user-specific image. This strategy provides a better initial setup for personalizing the diffusion models, effectively reducing required training steps. Some models~\cite{wei2023elite, chen2024subject,li2024blip} even support zero-shot personalization, although they underline that further fine-tuning can enhance personalization quality and address failure cases. Nonetheless, these methods are not viable for environments with severely limited computational resources, as they necessitate additional inference using large pre-trained models (e.g., 2.7B parameters for BLIP-2 in BLIP-Diffusion~\cite{li2024blip} and 2.5B for apprentice models in SuTI~\cite{chen2024subject}), which are substantially larger than standard diffusion models (e.g., 1B for Stable Diffusion v2~\cite{rombach2022high}), making their application challenging in on-device settings.

The second approach~\cite{kumari2023multi, han2023svdiff}, often involving LoRA, aims to reduce the number of updating parameters by limiting updates to specific layers or decomposing weight matrices. However, even with fewer parameters to update, these parameters reside within large pre-trained models, and thus the backward pass through the large models is required to compute gradients. Given limited computational resources, where even simple inference tasks with diffusion models can strain GPU memory, performing backpropagation while keeping the entire diffusion model in GPU memory remains a significant limitation.

A promising approach to address these challenges is side-tuning~\cite{zhang2020side,cai2020tinytl,zhang2022auxadapt,sung2022lst}, which fine-tunes a smaller auxiliary network rather than directly updating the parameters of a large pre-trained network. This method significantly reduces the heavy memory costs associated with computing backpropagation on the larger network. Particularly for Natural Language Processing (NLP) tasks, Ladder Side Tuning (LST)~\cite{sung2022lst} has proven effective, reducing the memory costs required for fine-tuning large language models (LLMs) by 69 percent. However, applying LST directly to diffusion U-Nets presents significant challenges. Unlike transformer layers in LLMs, which maintain consistent input and output dimensions, diffusion U-Nets have varying spatial dimensions and channels, as well as skip-connections across different blocks. Additionally, the requirements for structural pruning and weight initialization to build side-tuning networks further complicate the rapid adaptability of LST to personalization tasks across different subjects and domains.

\begin{figure}[t!]
\begin{center}
\includegraphics[width=\linewidth]{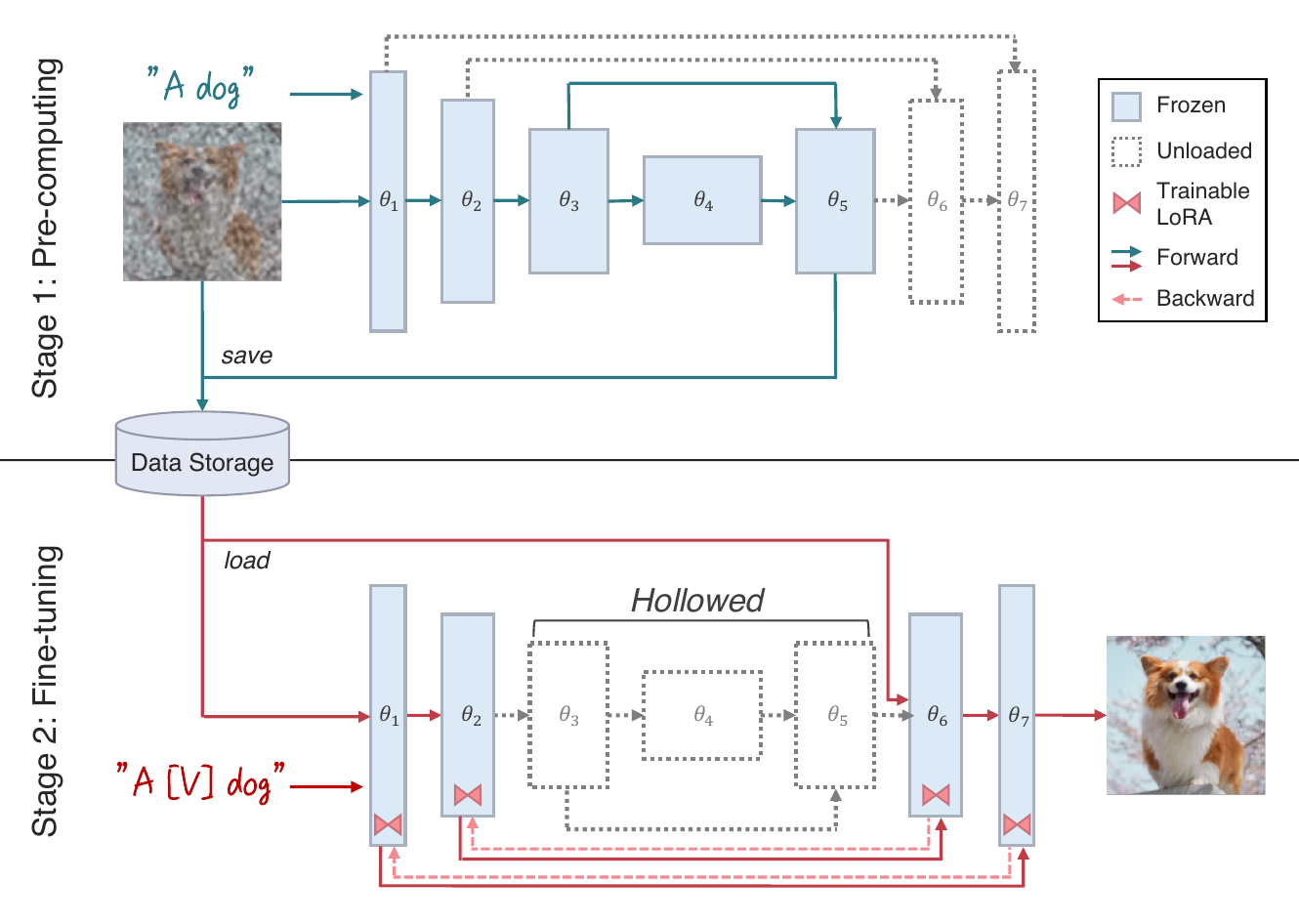}
\end{center}
   \caption{The LoRA personalization with Hollowed Net for resource-constrained environments. The input image is from the DreamBooth dataset~\cite{ruiz2023dreambooth}.}
\vspace{-10px}
\label{fig1}
\end{figure}

To this end, we introduce a novel personalization technique called \textit{Hollowed Net}, which is illustrated in Fig.~\ref{fig1}. Based on our observation that deep layers in the middle of diffusion U-Nets play significantly less important roles than the rest of the layers, we propose to fine-tune LoRA parameters for the personalization using Hollowed Net, a layer-pruned U-Net featuring a central hollow, which is constructed by temporarily removing the 
middle deep layers from the pre-trained diffusion U-Net. By utilizing the symmetrical "U-shape" architecture of the diffusion U-Net, we avoid complicated processes of applying structural pruning and weight initialization to build a side network, and neither additional models nor extensive pre-training with large datasets are required.

By fine-tuning LoRA parameters using Hollowed Net, we can significantly reduce the memory needed for storing model weights in GPU. Once the LoRA parameters are fine-tuned with Hollowed Net, they can be seamlessly transferred back to the original Diffusion U-Net for inference, without requiring any additional memory beyond the small set of transferred parameters. Our experiments demonstrate that Hollowed Net enables achieving performance that is comparable to or better than the direct fine-tuning with LoRA, while using 26 percent less GPU memory, which is only 11 percent increased GPU memory relative to an inference.

To the best of our knowledge, Hollowed Net is the first technique that addresses subject-driven generation in terms of memory efficiency. Our method shows how T2I diffusion models can be fine-tuned under extremely limited computational resources with as low GPU memory as required for inference. Furthermore, it is important to note that our method does not preclude the use of previously described strategies for efficient personalization. Both enhanced parameter-efficient strategies and improved initializations with additional pre-trained models can be integrated with our approach to further increase efficiency according to given resource constraints.

Our contributions can be summarized as follows:
\begin{itemize}
\item We introduce \textit{Hollowed Net}, a novel personalization technique for T2I diffusion models under limited computational resources. Our method significantly reduces the memory demands on GPU to levels as low as those required for inference, while maintaining a high-fidelity personalization capacity. This demonstrates its potential as a feasible on-device learning solution for resource-constrained devices.

\item Our method provides a scalable and controllable solution for on-device learning. As this method does not require any additional models or pre-training with large datasets, it is easily scalable to other architectures such as SDXL and Transformers. Moreover, we can simply adjust the fraction of hollowed layers to control the trade-offs between performance and memory requirements, depending on the target application and resources.

\item 
Unlike previous side-tuning methods, Hollowed Net does not need to be retained for inference. The LoRA parameters fine-tuned with Hollowed Net can be seamlessly transferred back to its original network, enabling inference with no additional memory cost.
\end{itemize}

\section{Related Works}

\subsection{Efficient Personalization of T2I Diffusion Models}
Recent research on the personalization of T2I diffusion models has introduced various methods to fine-tune the models for generating diverse images of user-specific subjects from a few given images. Two foundational works in this area are Textual Inversion and DreamBooth~\cite{gal2022image, ruiz2023dreambooth}. Textual Inversion~\cite{gal2022image} aims to learn new text embeddings to represent a given subject, while DreamBooth~\cite{ruiz2023dreambooth} proposes fine-tuning an entire diffusion model to align the subject with a unique token.

Building on these foundational works, recent research has focused on enhancing the efficiency of this personalization process, primarily through two approaches. The first approach involves decreasing the number of training steps, mostly by utilizing an additional large pre-trained model. A popular method is to use a pre-trained image/multi-modal encoder to generate personalized text embeddings or image prompts from a user-specific image~\cite{gal2023encoder, li2024blip, wei2023elite}. Other recent works~\cite{ruiz2023hyperdreambooth, chen2024subject} propose utilizing a set of pre-optimized LoRA parameters or millions of fine-tuned expert models to pre-initialize for efficient fine-tuning or enable zero-shot generation with in-context learning. While these models demonstrate significant reductions in the number of training steps, the requirement for additional large pre-trained models limits their application to on-device settings. Moreover, models with zero-shot personalization capacities~\cite{chen2024subject, li2024blip, wei2023elite} cannot be a one-size-fits-all solution for addressing different types of user-subject prompts. These models often struggle with flexibility in editing subjects or maintaining subject fidelity, and in these cases, additional fine-tuning with specific subjects is needed to further enhance their personalization capacity~\cite{chen2024subject, li2024blip}.

On the other hand, another stream of work adapts parameter-efficient fine-tuning (PEFT) approaches. These methods demonstrate significant reductions in the number of training parameters by limiting updates to a small subset of model weights in cross-attention layers~\cite{kumari2023multi} or further reducing the updating parameters by applying singular vector decomposition to weight matrices~\cite{han2023svdiff}. However, these methods are still limited in environments with extremely low computational resources, as they require backpropagation over large diffusion models and do not reduce memory usage from the model weights. Therefore, it is crucial to explore new approaches for personalizing T2I diffusion models in resource-limited settings, as we propose with our novel method, Hollowed Net. Notably, our method can be integrated with previously discussed techniques to further improve efficiency based on specific resource constraints.

\subsection{Fine-Tuning with Side Networks}
The idea of of side-tuning has been introduced by Zhang et al.~\cite{zhang2020side}, proposing the training of a lightweight "side" network instead of directly fine-tuning a pre-trained network for adaptation. In terms of efficiency, Cai et al.~\cite{cai2020tinytl} has demonstrated an additional lightweight residual module can reduce memory overhead associated with the activations of the original network. Similarly, AuxAdapt~\cite{zhang2022auxadapt} has shown that a small auxiliary network can be fine-tuned to adjust the main network's decisions, enabling efficient test-time adaptation for video semantic segmentation tasks.

In the context of generative models, LST~\cite{sung2022lst} has demonstrated the effectiveness of side networks for different NLP tasks with LLMs by introducing a small side network that takes intermediate activations of the main network as input via shortcut connections. However, directly applying LST to diffusion U-Nets poses challenges due to varying spatial dimensions, channel sizes, and skip-connections across blocks, unlike the consistent dimensions in transformer layers of LLMs. Furthermore, the structural pruning and specific weight initialization required to construct side-tuning networks complicate LST's adaptability for personalized tasks across a range of subjects and domains.

\subsection{Layer Pruning of Large Generative Models}
Several concurrent works demonstrate that layer-pruning methods can be applied to generative models, particularly for NLP tasks. Gromov et al.~\cite{gromov2024unreasonable} suggest that for fine-tuning LLM models, up to 40\% of deep layers can be removed, while still achieving comparable results. The authors propose that the optimal block of layers to prune can be selected based on similarity across layers. Similarly, Kim et al.~\cite{kim2024shortened} also propose a depth-pruning approach by evaluating block-level importance.

These approaches differ from ours due to the distinct characteristics of LLMs versus diffusion U-Nets. The aforementioned approaches involve the complete removal of deep layers for both fine-tuning and inference, considering that those layers store less critical knowledge. However, our study finds that the deep layers of diffusion U-Nets may be less involved with personalization but still contain crucial high-level image features for generating high-fidelity images. Thus, their removal can lead to severe performance degradation, even with additional pre-training~\cite{kim2023bk}, as shown in Appendix~\ref{pruning}. This highlights the importance of our two-stage fine-tuning strategy, which excludes layers during fine-tuning to reduce memory overhead while preserving the knowledge from these excluded layers throughout both training and inference stages.

\section{Preliminaries} 
In this section, we describe some preliminaries on T2I diffusion models. First, we discuss the basics of Stable Diffusion (SD) model and how they can be used for fine-tuning. The SD model is a large foundational T2I model, pre-trained on large amount of text-image pairs ${(P,x)}$, where we have image $x$ and associated text prompt $P$. The SD contains the following components: (a) Autoencoder consisting of the encoder-decoder pair $(\mathcal{E}, \mathcal{D})$, (b) Text Encoder as CLIP $E_{T}(\cdot)$, and (c) Conditional Diffusion Model as U-Net $\epsilon_{\theta}(\cdot)$. The encoder $\mathcal{E}(\cdot)$ processes an image $x$ into a latent space $z=\mathcal{E}(x)$, and the decoder is used to reconstruct the input image from latent $z$ such that $x \approx \mathcal{D}(z)$. The diffusion process of SD is conducted in the latent space. For a randomly sampled noise $\epsilon \sim \mathcal{N}(0,I)$ at time step $t$, the standard scheduler produces a noisy latent code $z_{t} = \alpha_t z + \sigma_t \epsilon$, where $\alpha_t$ and $\sigma_t$ are coefficients controlling the noise schedule. The conditional diffusion model $\epsilon_{\theta}$ is trained using the following de-noising objective:
\begin{equation}
\label{eqn:ft}
\texttt{min} \quad \mathbb{E}_{P,z,\epsilon,t}[||\epsilon - \epsilon_\theta (z_t, t, E_T(P))||_2^2].
\end{equation}

After the training is carried out, the conditioned model $\epsilon_{\theta}(\cdot)$ is used to predict the noise by using the conditional embedding $E_{T}(P)$ and time step $t$ as input. 
To personalize diffusion models for subject-driven generation introduced by~\cite{ruiz2023dreambooth}, the same loss is used except that the data is sampled from user-specific subjects such as dog, person, backpack, and etc. For the prompt, a special identifier $S*$ is used and described as "a $S*$ person", "a $S*$ backpack", etc. For regularization, \cite{ruiz2023dreambooth} introduces an additional class-specific prior preservation loss term, written as 
\begin{equation}
\label{eqn:db}
\texttt{min} \quad \mathbb{E}_{z,\epsilon,t}[||\epsilon_{pr} - \epsilon_\theta (z'_t, t, E_T(P_{pr}))||_2^2]
\end{equation}
where $\epsilon_{pr}$ is the ground truth noise for the data generated using the frozen pre-trained diffusion model with prompts $P_{pr}$ described more generic as "a person", "a backpack", and etc.

The diffusion U-Net can be fully fine-tuned, but it is also possible to fine-tune only a subset of parameters with LoRA~\cite{hu2021lora} for better efficiency. In LoRA, network weight residuals $\Delta W$ are fine-tuned instead of the full weights $W$. For the fine-tuning of $\Delta W$, it is further decomposed into low-rank matrices $A$ and $B$ such that $\Delta W = AB$. Since $A$ and $B$ are low-rank matrices, the total number of parameters to optimize in $\Delta W$ is significantly smaller than in $W$.

\begin{figure}[!tbp]
  \centering
  \subfloat[DreamBooth]{\includegraphics[width=0.5\textwidth]{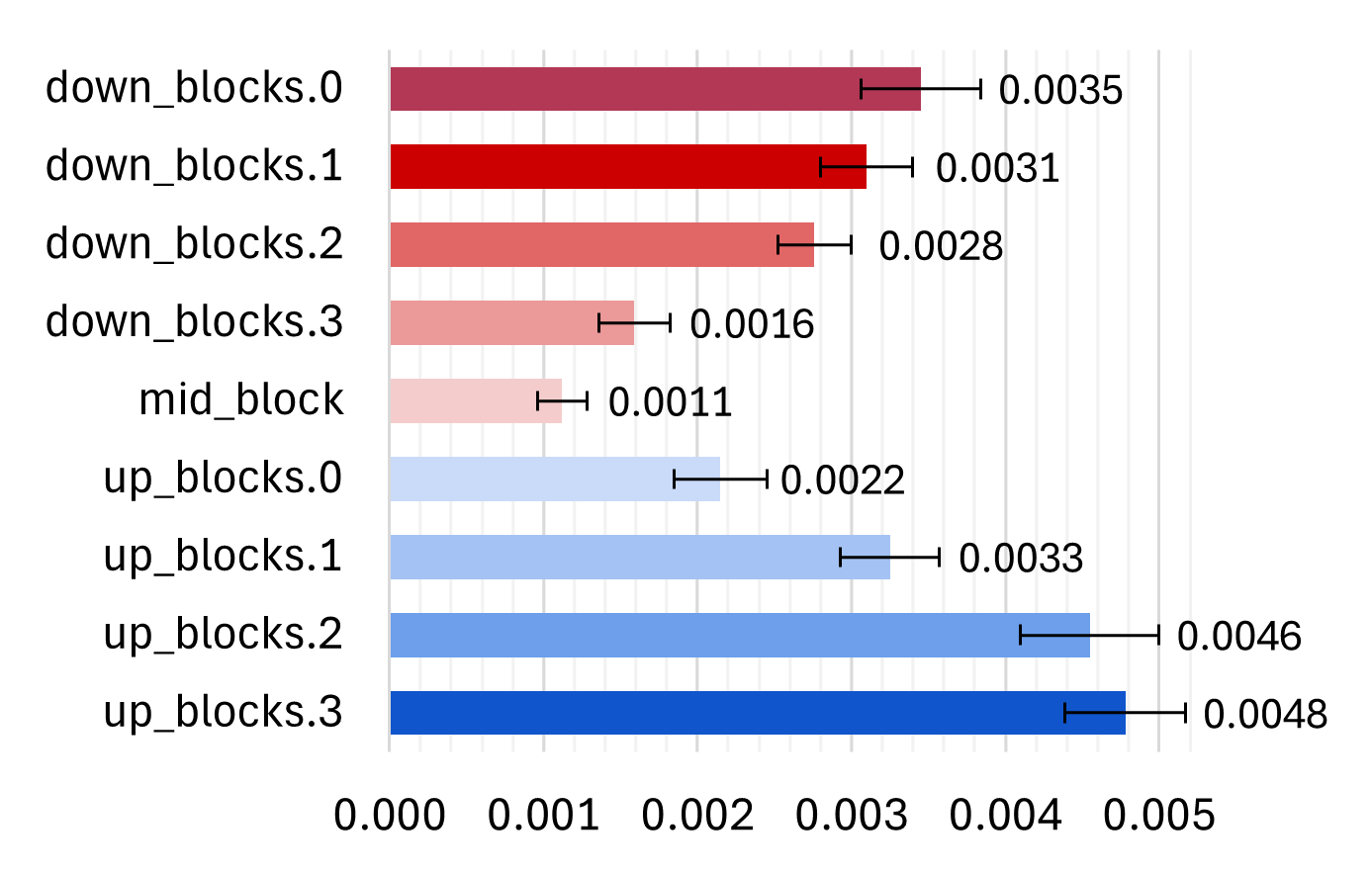}}
  \hfill
  \subfloat[CustomConcept101]{\includegraphics[width=0.5\textwidth]{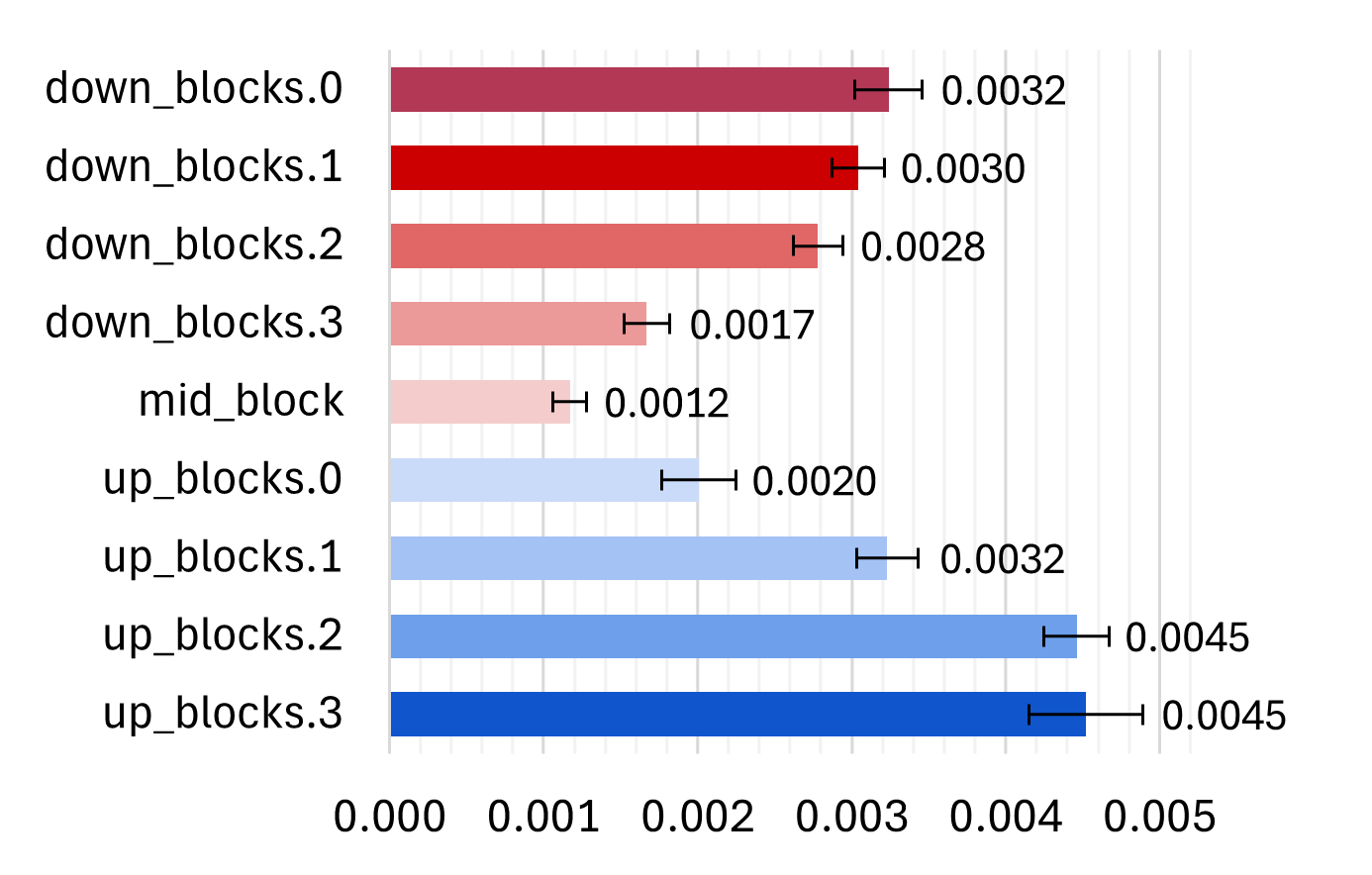}}
  \caption{Analysis of the LoRA weight change before and after personalization, per block of U-Net.}
\label{fig:observation}
\vspace{-10px}
\end{figure}

\section{Methodology}

In this section, we describe the details of our novel memory-efficient personalization technique, Hollowed Net, and its fine-tuning strategy. We begin by identifying less significant layers for personalization from diffusion U-Nets. Based on these observations, we explain how to construct Hollowed Net from a pre-trained U-Net. Next, we present our fine-tuning and inference processes for memory-efficient personalization of T2I diffusion models.

\subsection{Analysis of the LoRA Weight Changes per Block of U-Net}

To achieve the goal of reducing the required memory for fine-tuning a diffusion model, we first identify less significant layers in the diffusion U-Net for personalization. Similar to Li et al.~\cite{li2020few}, Kumari et al.~\cite{kumari2023multi} and Shah et al.~\cite{shah2023ziplora}, we analyze the LoRA weight changes $\Delta W$ in the fine-tuned model for each block:
\vspace{-2px}
\begin{equation}
\Delta W = \frac{1}{n} \sum_{i=1}^{n} |w_i - w'_i|,
\end{equation}

\noindent where 
$w$ and $w'$ respectively represent the weights before and after personalization,
and $n$ is the total number of weights in a specific block. This represents the average weight change per element $i$. Figure \ref{fig:observation} shows the analysis of the weight changes $\Delta W$ before and after personalization for each block of U-Net: (a) for all subjects from the DreamBooth dataset and (b) for all subjects from the CustomConcept101 dataset by fine-tuning Stable Diffusion v2.1 diffusion model~\cite{rombach2022high} for 1000 steps with a learning rate of 1e-4. The x-axis shows the changes in LoRA weights before and after personalization, while the y-axis of each plot represents the specific U-Net blocks. For each dataset, we average the weight changes across subjects and provide error bars to indicate the statistical variance within each dataset.

From the figures, we observe that the average weight changes tend to be close to zero around the central blocks and become increasing for the layers farther from the ${\rm mid}\_{\rm block}$. This demonstrates that the blocks around the center are less involved in the personalization compared to those at the beginning and end of the U-Net (e.g., ${\rm down}\_{\rm blocks.0}$, ${\rm down}\_{\rm blocks.1}$, ${\rm up}\_{\rm blocks.2}$, and ${\rm up}\_{\rm blocks.3}$). We leverage this characteristic for designing Hollowed Net.

\subsection{Hollowed Net}

Based on the aforementioned observations, we propose fine-tuning a layer-pruned U-Net, which we refer to as Hollowed Net, instead of directly fine-tuning the entire diffusion model. The core concept of Hollowed Net involves removing deep layers that are not vital for personalization from a pre-trained diffusion U-Net. This strategy decreases the need to store the entire model in GPU memory, thereby reducing the memory cost associated with the model's weights.

However, unlike transformer layers in large language models, where input and output maintain the same data structure, the alterations in spatial and channel dimensions in U-Net architectures complicate the removal of its deep layers in the middle. To address this, we utilize the symmetrical "U-shape" architecture of the diffusion U-Net, where each down-block layer's output is concatenated with a corresponding up-block layer's input via a skip-connection. This design permits us to select any up-block layer skip-connected to a down-block layer and hollow out the middle layers between the pair, ensuring that the processed information from the remaining down-blocks can still be transferred to the remaining up-blocks without the need for additional projection layers to adjust for dimensional differences. The missing input for the upper layer, due to the removal of the middle layers, is replaced with the pre-computed output from the full diffusion U-Net, which is illustrated in the next section.

\begin{figure}[t!]
\begin{center}
\includegraphics[width=\linewidth]{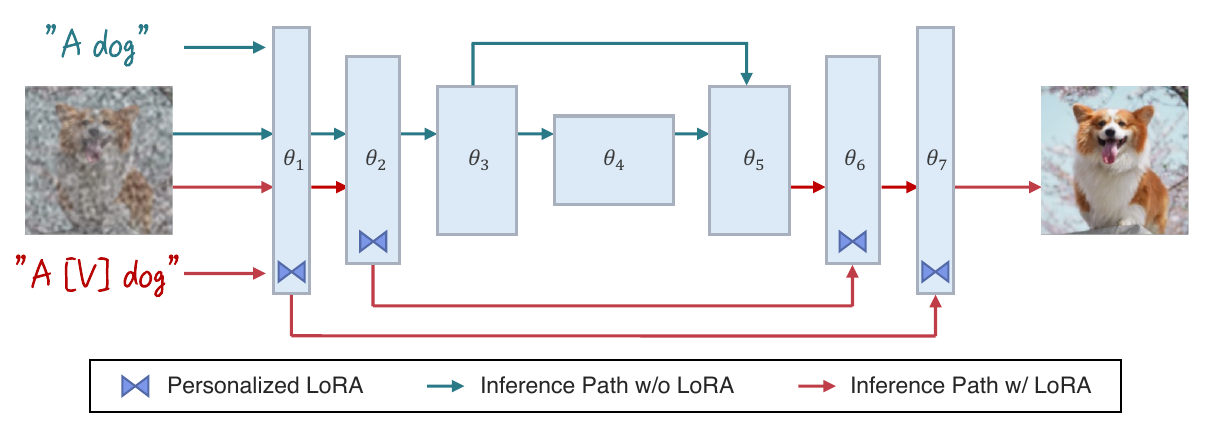}
\end{center}
   \caption{The inference process with personalized LoRA parameters transferred from Hollowed Net to the original U-Net. The input image is from the DreamBooth dataset~\cite{ruiz2023dreambooth}.}
\label{fig2}
\vspace{-10px}
\end{figure}

\subsection{LoRA Personalization with Hollowed Net}

To optimize GPU memory utilization, we propose a two-stage fine-tuning strategy: (1) pre-computing intermediate activations of the original diffusion U-Net and (2) fine-tuning the Hollowed Net using the pre-computed activations, as shown in the upper and bottom half of Fig.~\ref{fig1}, respectively. Initially, we conduct a forward pass with a pre-trained diffusion model for the specified number of pre-computing steps. During each step, given input images and sampled noise, we calculate and store intermediate activations in the data storage, which serve as inputs for the upper-block layer of the Hollowed Net. We also store the sampled noises, time steps, and the IDs when there are multiple user images.

Once the data from the pre-trained model is pre-computed, we fine-tune the Hollowed Net by loading data from data storage, thereby avoiding the need to keep the original model in GPU memory. To further improve efficiency, we apply LoRA fine-tuning for the Hollowed Net instead of updating entire parameters. The reduced number of parameters of the Hollowed Net decreases the required GPU memory, satisfying the device's low memory I/O threshold and computational load during backpropagation.

Additionally, our inference process ensures that both the original diffusion model and Hollowed Net are not simultaneously maintained on GPU. Unlike side-tuning networks~\cite{zhang2020side,zhang2022auxadapt,sung2022lst} that differ in architecture and parameters from their original models, Hollowed Net maintains the same architectures and parameters as the original diffusion U-Net, except for the removed middle layers. Thus, the LoRA parameters fine-tuned on Hollowed Net can be seamlessly transferred to the corresponding layers in the original U-Net. As depicted in Fig.~\ref{fig2}, there are two inference paths, respectively corresponds to each stage of fine-tuning. The first path (green line) represents the process of computing intermediate activations without using LoRA, aligning with the pre-computing stage. The second path (red line) involves using personalized LoRA parameters, which matches the application of these parameters for generating personalized images during the fine-tuning stage. By sequentially executing these paths, we enable personalized generation using the transferred LoRA parameters without requiring additional memory beyond the small set of LoRA parameters.

\begin{table}[t!]
\caption{The quantitative comparisons of fine-tuning methods with three evaluation metrics. The number of parameters are the ones held in GPU memory during fine-tuning stage. The results are obtained by averaging over four runs with different seeds (standard deviation is added in a small-sized text).}
\label{table1}
\vspace{8pt}
\centering

\resizebox{\textwidth}{!}{
\begin{tabular}{cccccccccccccc}
\toprule
\multirow{2}{*}{Method}          & \multicolumn{2}{c}{\# of Parameters}              &  & \multicolumn{2}{c}{Training Memory}                          &  & \multicolumn{3}{c}{DreamBooth}                                                     &  & \multicolumn{3}{c}{CustomConcept101}                                               \\ \cmidrule(lr){2-3} \cmidrule(lr){5-6} \cmidrule(lr){8-10} \cmidrule(lr){12-14} 
                                 & Base                 & LoRA                 &  & \multicolumn{1}{c}{Peak} & \multicolumn{1}{c}{Comp. w/ Inf.} &  & \multicolumn{1}{c}{DINO} & \multicolumn{1}{c}{CLIP-I} & \multicolumn{1}{c}{CLIP-T} &  & \multicolumn{1}{c}{DINO} & \multicolumn{1}{c}{CLIP-I} & \multicolumn{1}{c}{CLIP-T} \\ \midrule

\raisebox{0.4ex}{Full FT}      & \raisebox{0.37ex}{866M}                 & \raisebox{0.37ex}{-}                    &  & \raisebox{0.37ex}{16.62GB}                  & \raisebox{0.37ex}{+376\%}                          &  & \raisebox{-0.5ex}{\shortstack{0.663 \\ \scriptsize{$\pm0.013$}}} & \raisebox{-0.5ex}{\shortstack{0.802 \\ \scriptsize{$\pm0.007$}}}   & \raisebox{-0.5ex}{\shortstack{0.302 \\ \scriptsize{$\pm0.002$}}}   &  & \raisebox{-0.5ex}{\shortstack{0.605 \\ \scriptsize{$\pm0.005$}}} & \raisebox{-0.5ex}{\shortstack{0.773 \\ \scriptsize{$\pm0.006$}}}   & \raisebox{-0.5ex}{\shortstack{0.302 \\ \scriptsize{$\pm0.002$}}}   \\ \midrule

\shortstack{LoRA FT \\ (r=128)} & \raisebox{1ex}{866M}                 & \raisebox{1ex}{27M}                  &  & \raisebox{1ex}{5.23GB}                   & \raisebox{1ex}{+50\%}                          &  & \shortstack{0.658 \\ \scriptsize{$\pm0.001$}}  & \shortstack{0.806 \\ \scriptsize{$\pm0.005$}}   & \shortstack{0.299 \\ \scriptsize{$\pm0.002$}}   &  & \shortstack{0.603 \\ \scriptsize{$\pm0.008$}} & \shortstack{0.773 \\ \scriptsize{$\pm0.005$}}   & \shortstack{0.302 \\ \scriptsize{$\pm0.002$}}   
\\ 

\shortstack{LoRA FT \\ (r=1)}    & \raisebox{1ex}{866M}                 & \raisebox{1ex}{207K}                 &  & \raisebox{1ex}{4.84GB}                   & \raisebox{1ex}{+39\%}                          &  & \shortstack{0.516 \\ \scriptsize{$\pm0.011$}}  & \shortstack{0.738 \\ \scriptsize{$\pm0.003$}}   & \shortstack{0.314 \\ \scriptsize{$\pm0.001$}}   &  & \shortstack{0.522 \\ \scriptsize{$\pm0.008$}} & \shortstack{0.737 \\ \scriptsize{$\pm0.005$}}   & \shortstack{0.305 \\ \scriptsize{$\pm0.001$}}   \\
\midrule
\shortstack{\textbf{Hollowed Net}\\ (Ours)} & \raisebox{1ex}{\textbf{527M}}                 & \raisebox{1ex}{\textbf{24M}}                  &  & \raisebox{1ex}{\textbf{3.88GB}}          & \raisebox{1ex}{\textbf{+11\%}}                 &  & \shortstack{0.660 \\ \scriptsize{$\pm0.011$}} & \shortstack{0.805 \\ \scriptsize{$\pm0.006$}}   & \shortstack{0.300 \\ \scriptsize{$\pm0.001$}}   &  & \shortstack{0.603 \\ \scriptsize{$\pm0.007$}} & \shortstack{0.773 \\ \scriptsize{$\pm0.005$}}   & \shortstack{0.302 \\ \scriptsize{$\pm0.002$}}   \\ 
\bottomrule
\end{tabular}

}
\end{table}

\begin{table}
\centering
\begin{minipage}[t]{0.415\textwidth}  

    \centering
    \caption{Human evaluation results}
    \vspace{5pt}
    \resizebox{\textwidth}{!}{ 
\begin{tabular}{ccc}
\toprule
Method & Subject Fidelity & Text Fidelity \\ \midrule
Hollowed Net     & 31.2\%                    & 18.1\%                 \\ 
Tie              & 49.3\%                    & 69.4\%                 \\ 
LoRA FT          & 19.5\%                    & 12.5\%                 \\ \bottomrule
\end{tabular}
\label{human_eval}
}
\end{minipage}%
\hspace{16pt}
\begin{minipage}[t]{0.48\textwidth}  
    \centering
    \caption{Computational loads (FLOPs)}
    \vspace{5pt}
    \resizebox{\textwidth}{!}{ 
    \begin{tabular}{cccc}
    \toprule
    Method         & Pre-computing & Fine-tuning & Inference\\ \midrule
    Hollowed Net & 0.238T                & 2.004T        &0.920T        \\
    LoRA FT   &     -           & 2.148T     & 0.716T     \\ \bottomrule
    \end{tabular}
\label{flops}
    }
\end{minipage}
\vspace{-10px}
\end{table}

\section{Experiments}
\subsection{Experimental Settings}
We conduct experiments following the protocol proposed in DreamBooth~\cite{ruiz2023dreambooth}. We use a total of 131 subjects for experiments, utilizing both the DreamBooth~\cite{ruiz2023dreambooth} and CustomConcept101~\cite{kumari2023multi} datasets. The DreamBooth dataset includes 30 image sets from 15 different classes, each containing 4-6 images of a given subject. The subjects are divided into living subjects and objects, and 25 different prompts are assigned based on this division. Meanwhile, the CustomConcept101 dataset includes 101 image sets, each containing 3-15 images of a given subject. The subjects consist of 15 different large categories, with 20 unique prompts assigned to each category. For evaluation, four images with different fixed random seeds are generated per subject per prompt for both datasets.

We adopt the three evaluation metrics from~\cite{ruiz2023dreambooth}: DINO and CLIP-I for subject fidelity and CLIP-T for prompt fidelity. DINO and CLIP-I are the average pairwise cosine similarities between
feature embeddings of the real and generated images, using DINO ViT-S/16 and CLIP ViT-B/32, respectively. As DINO is more sensitive to differences between subjects of the same class due to its training on instance discrimination, the DINO score is considered the preferred metric for measuring subject fidelity. The CLIP-T score is the average cosine similarity between text prompt embeddings and image CLIP embeddings. We use the Stable Diffusion v2.1 diffusion model~\cite{rombach2022high}. Following DreamBooth~\cite{ruiz2023dreambooth}, we use a prior preservation loss with $\sim$1000 pre-generated class samples. LoRA~\cite{hu2021lora} is applied for the cross and self-attention layers and fine-tuned for $\sim$1000 steps. We use AdamW optimizer with the learning rate of 1e-5 for full-finetuning and 1e-4 for the others. Assuming a resource-constrained environment, we use a batch size of 1 and do not update the pre-trained text encoder, while text embeddings are pre-computed before fine-tuning.

\begin{figure}[t!]

\begin{center}
\includegraphics[width=\linewidth]{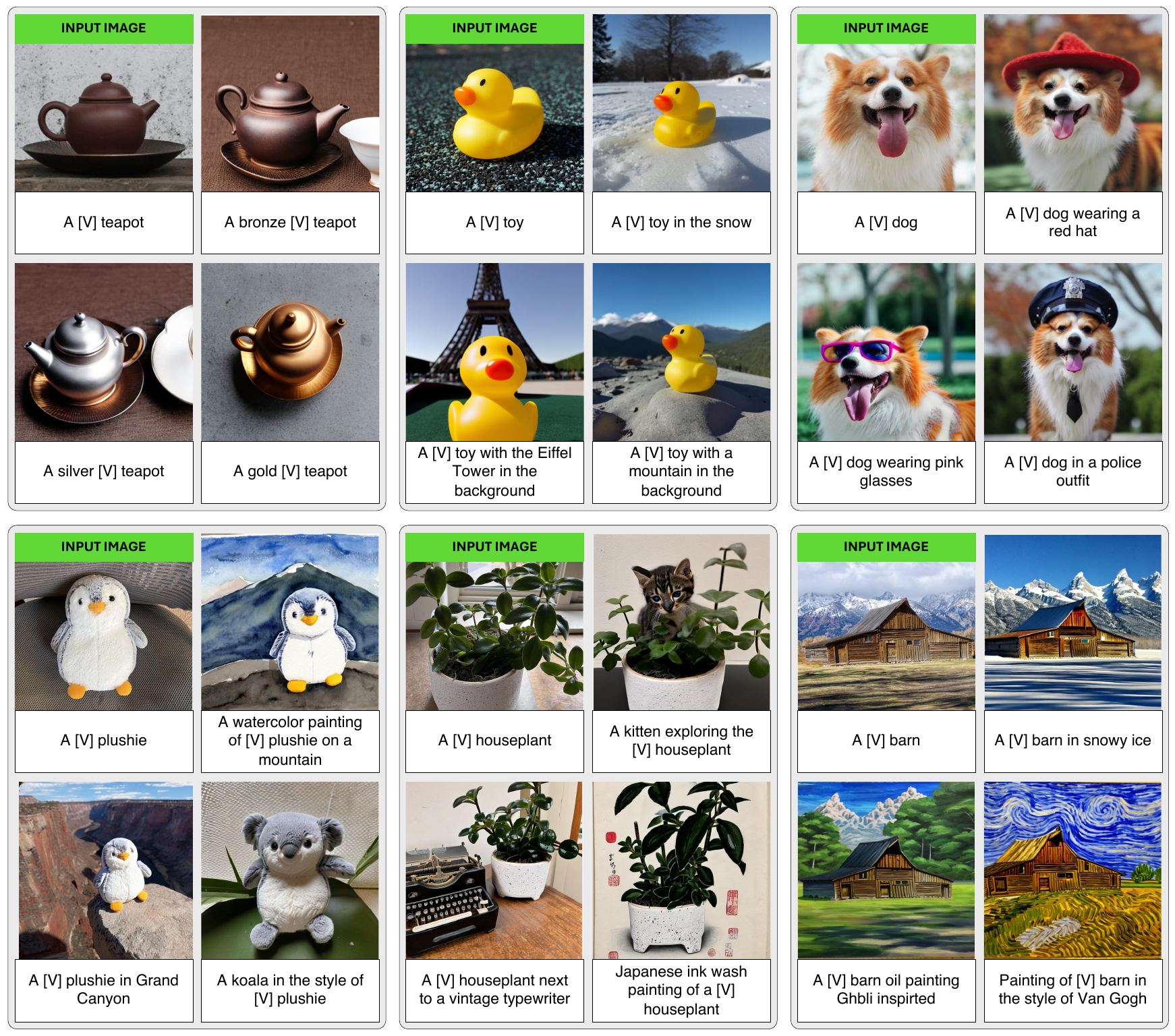}
\end{center}
   \caption{Qualitative generation results of Hollowed Net with different subjects and prompts. The upper half are the examples from the DreamBooth dataset~\cite{ruiz2023dreambooth}, and the lower half are the examples from the CustomConcept101 dataset~\cite{kumari2023multi}.}

\vspace{-15px}
\label{fig4}
\end{figure}

\subsection{Results}
\label{results}
In this section, we present the results of our proposed Hollowed Net to evaluate its effectiveness in terms of both memory efficiency and personalization performance. We conduct experiments with Hollowed Net, applying a hollowed fraction of 39.2\%. Architectural details are provided in Appendix~\ref{appendix_a}. Ablation studies on different fractions of hollowed layers can be found in Sec.~\ref{ablation}. In the main results, the rank of Hollowed Net is fixed to $128$. Experimental results on different ranks are presented in Appendix~\ref{appendix_b}.

\paragraph{Quantitative Evaluation}
The quantitative results are displayed in Table~\ref{table1}. For comparison baselines, we implement full fine-tuning (Full FT) and LoRA fine-tuning (LoRA FT) methods with rank 128 and rank 1~\cite{hu2021lora}. We find that while Full FT results in slightly higher performance than other methods, particularly in terms of DINO, the differences between Full FT and Hollowed Net are not significant as it is within the range of standard deviations of Full FT results across different seeds. Moreover, Full FT requires more than 16GB of GPU memory which is nearly 4.7 times the memory cost of performing an inference with a diffusion U-Net. Clearly, this is not a feasible solution for on-device learning, where computation resources, especially memory I/O, are extremely limited.

Our Hollowed Net demonstrates its superior memory efficiency based on a significant reduction in model size, requiring only 3.88GB of GPU memory usage for fine-tuning. This is only an 11\% increase compared to inference. Its personalization performance is comparable to or marginally better than that of LoRA fine-tuning using the same rank ($r=128$), while LoRA requires a 50\% increase in GPU memory compared to inference. Using the lowest rank of LoRA ($r=1$) does not compete with Hollowed Net either, as its memory efficiency is limited by the need to run backpropagation on the entire U-Net, even though the number of fine-tuning parameters is significantly small. Additionally, the use of a low number of fine-tuning parameters significantly degrades personalization capacity.

For human evaluation, we conduct user studies with 40 participants, each completing a set of 25 comparative tasks. In each task, participants are presented with a reference image, a prompt, and two generated images (A and B). They answer two questions: subject fidelity and text fidelity. Each pair of generated images, A and B, is created using Hollowed Net and LoRA FT, and the labels (A or B) are randomly assigned for each task. Table~\ref{human_eval} displays the results of these user studies. These findings confirm that users generally perceive the images generated by Hollowed Net and LoRA FT to be similar in both subject fidelity and text fidelity, consistent with the main results presented in Table~\ref{table1}.

Additionally, we include the analysis of computational loads for Hollowed Net and LoRA FT in Table~\ref{flops}. Each number corresponds to one step of each stage: one forward pass for pre-computing and inference and one forward+backward pass for fine-tuning. For the fine-tuning of Hollowed Net, $\sim$1000 steps are required, totaling $2.004 \times 1000 = 2004$ TFLOPs. For pre-computing, we find 200 pre-computed samples are sufficient to achieve high-fidelity results (see Appendix \ref{appendixC} for a detailed analysis), requiring $0.238 \times 200 = 47.6$ TFLOPs of additional computation. Therefore, the total computation required for training with Hollowed Net is $2004+47.6=2051.6$ TFLOPs, which is lower than $2.148 \times 1000 = 2148$ TFLOPs needed for LoRA FT. On the other hand, for running an inference pass, Hollowet Net requires approximately $0.204$ TFLOPs more than LoRA FT, as it needs to repeat part of the early down-blocks to reproduce the path used in training.

\begin{figure}[t!]

\begin{center}
\includegraphics[width=1.0\linewidth]{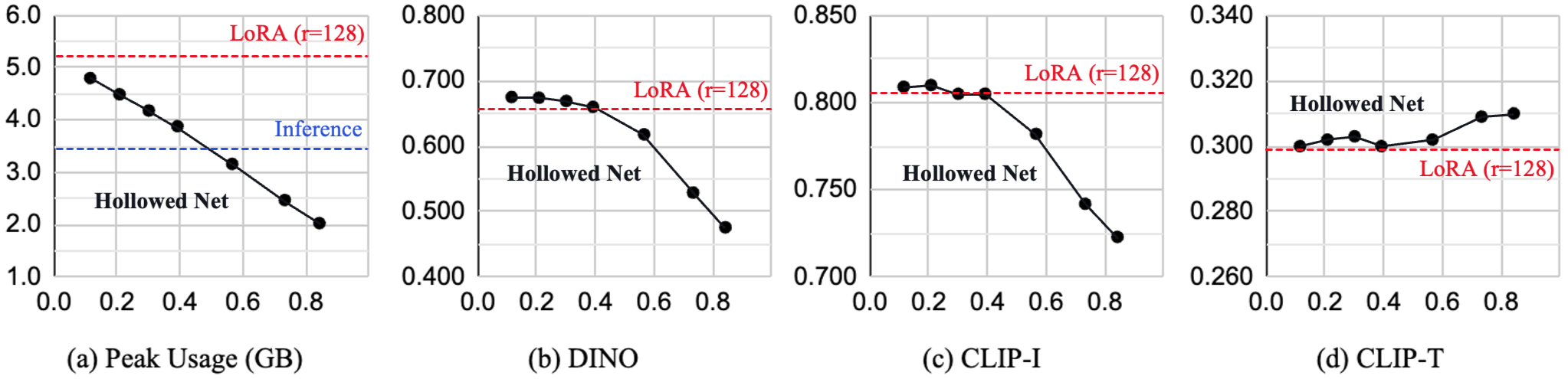}
\end{center}
   \caption{Analysis of different fractions of hollowed layers. For all figures, the x-axis represents the fractions of layers removed from the pre-trained diffusion U-Net. The y-axis corresponds to the metric used for each figure.}
\vspace{-10px}
\label{fig5}
\end{figure}

\begin{figure}[t!]
\begin{center}
\includegraphics[width=1.0\linewidth]{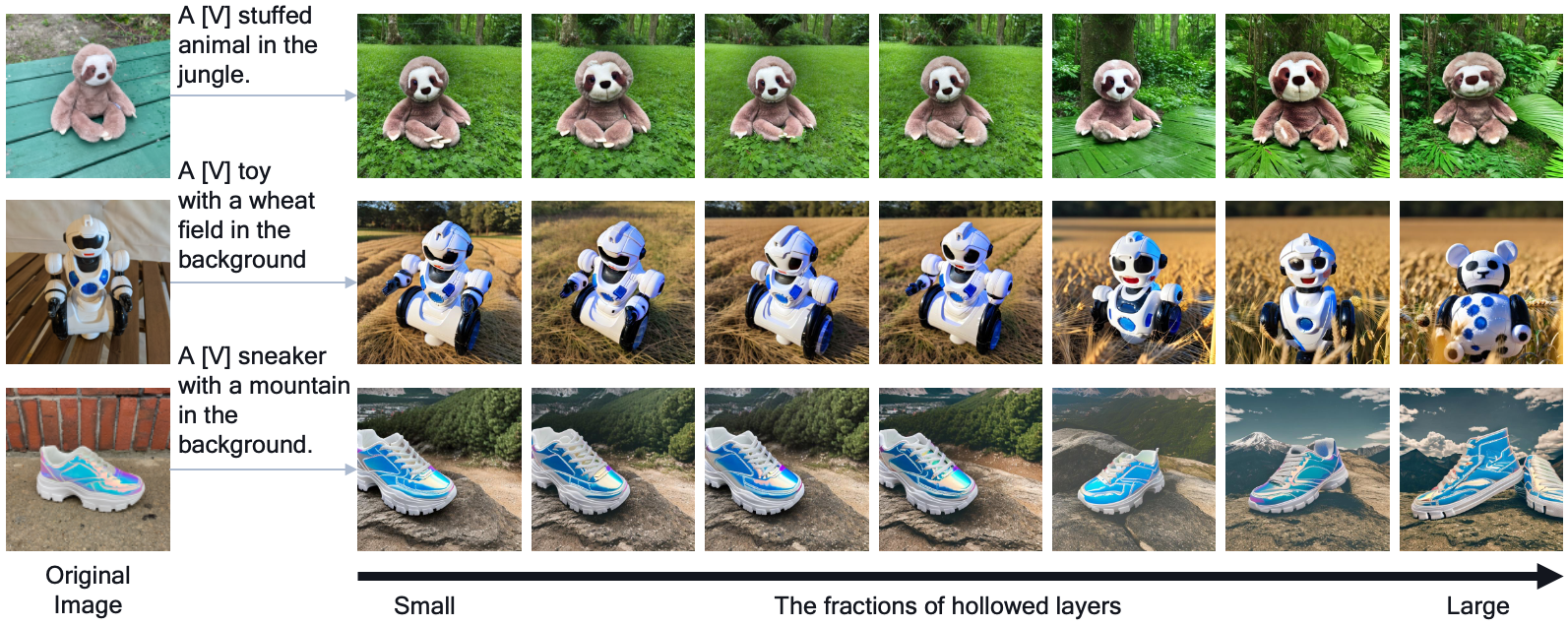}
\end{center}
   \caption{Qualitative results with different fractions of hollowed layers given three subjects from the DreamBooth dataset~\cite{ruiz2023dreambooth}.}
   \label{fig_Q_frac}
\vspace{-5px}
\end{figure}

\paragraph{Qualitative Evaluation}
In Fig.~\ref{fig4}, we present qualitative generation results of Hollowed Net for various subjects and prompts. The upper half shows examples from the DreamBooth dataset, and the lower half displays examples from the CustomConcept101 dataset. These results demonstrate that Hollowed Net effectively captures the visual details of the target subjects, while maintaining high text-image alignment for different types of applications including property modification, recontextualization, accessorization, and artistic rendition. Its ability enabling high-fidelity personalization with memory costs as low as those of inference makes it an efficient solution for a range of on-device applications with constrained computational resources. Additional qualitative examples with SDXL~\cite{podell2023sdxl} are included in Appendix~\ref{appendix_d}, illustrating the scalability of our approach in a larger model.

\subsection{Ablation Study on Fractions of Hollowed Layers}
\label{ablation}
Based on the symmetrical "U-shape" architecture of the diffusion U-Net, we can design different Hollowed Net architectures by selecting a different up-block layer skip-connected to a down-block layer and hollowing out the middle layers between the pair. Figure~\ref{fig5} presents experimental results across different fractions of hollowed layers, ranging from around 10\% to 85\% of layers removed. In Fig.~\ref{fig5}, we observe the peak GPU memory usage decreasing linearly with layer removal, as fewer model weights need to be stored on the GPU during backpropagation. Analyzing the DINO and CLIP-I scores in Fig.~\ref{fig5} (b) and (c), we find that the model's capacity to preserve subject fidelity remains comparable to or slightly better than LoRA until around 39.2\% of layers are removed, where memory cost reduces nearly to the level of inference. Beyond this threshold, however, subject fidelity significantly diminishes, as fewer layers essential for personalization are included in the Hollowed Net. This effect of hollowed layer fractions is also visible in the qualitative results in Fig.~\ref{fig_Q_frac}. Meanwhile, the CLIP-T score does not exhibit a general trend, except in cases of very high hollowed fractions, where the model is not capable of personalization, and thus generates images solely based on a given prompt. However, note that the increase in CLIP-T remains marginal, as Hollowed Nets with low hollowed fractions also maintain a high capacity for text-image alignment.

\section{Conclusion}
In conclusion, our paper introduces a novel approach for on-device personalization through memory-efficient fine-tuning with Hollowed Net. Hollowed Net effectively leverages the architecture of the diffusion U-Net, enabling fine-tuning with significantly reduced memory costs by minimizing the model's size during fine-tuning without requiring any additional processes such as structural pruning or pre-training on large-scale datasets. However, we observe that, due to the use of non-personalized prompts with the original network, the model's performance can be sensitive to the granularity of class token definitions. For example, the DreamBooth dataset contains "poop emoji" images, for which the class token is very coarsely defined as "toy". In this case, non-personalized intermediate activations generated with prompts using "toy" struggle to effectively correlate and generate "poop emoji" image. Therefore, a careful choice of fine-grained class tokens is necessary for the effective application of Hollowed Net.

Additionally, it is worth noting that our methodology is orthogonal to existing different PEFT methods~\cite{dettmers2024qlora, liu2024dora} and quantization methods~\cite{shang2023post, li2023q}. Thus, our approach offers substantial potential for further memory reduction, which is crucial for training under constrained computational resources. Furthermore, while our primary focus in this paper has been on image generation tasks, our method is not limited to diffusion models and can be seamlessly extended to various NLP tasks with LLMs, which we leave for future work. We anticipate that Hollowed Net will be applied to a wide range of tasks requiring constrained computational resources, serving as an efficient solution for various on-device applications. 

\newpage
{\small
\bibliographystyle{unsrt}
\bibliography{main}
}

\newpage

\appendix

\section{Experiments with Layer-Pruned Diffusion Models}
\label{pruning}

As shown in recent work~\cite{kim2023bk}, layer pruning involves the complete removal of selected layers, which necessitates extensive pre-training on large datasets to recover lost information and restore model functionality. However, diffusion models often suffer from substantial performance degradation post-pruning, as the lost information may not be fully recoverable through pre-training.

Table~\ref{bk-sdm} presents experiments with BK-SDM~\cite{kim2023bk} models, layer-pruned SD models, using rank-128 LoRA. Compared to the results in Table~\ref{table1}, these models achieve memory usage comparable to Hollowed Net but show significant performance degradation. Despite extensive pre-training, their performance remains compromised.

In contrast, Hollowed Net does not completely remove deep layers and requires no additional pre-training. Instead, we temporarily exclude selected layers during fine-tuning while preserving essential information through a pre-computation stage. Notably, despite this added stage, the overall computational load for training Hollowed Net can remain more efficient than LoRA fine-tuning, as discussed in Sec.~\ref{results} and Appendix~\ref{appendixC}.

\begin{table}[h!]
\centering
\caption{Quantitative results of BK-SDM on the DreamBooth dataset~\cite{ruiz2023dreambooth}.}
\label{bk-sdm}
\resizebox{\textwidth}{!}{
\begin{tabular}{lccccc}
\toprule
Method & \# of Parameters & Training Memory & DINO & CLIP-I & CLIP-T \\
\midrule
BK-SDM-Base  & 595.7M & 3.546GB & 0.629 $\pm$ 0.012 & 0.788 $\pm$ 0.007 & 0.300 $\pm$ 0.001 \\
BK-SDM-Small & 496.2M & 3.133GB & 0.602 $\pm$ 0.013 & 0.774 $\pm$ 0.008 & 0.298 $\pm$ 0.001 \\
\bottomrule
\end{tabular}
}
\end{table}

\begin{figure}[h!]
\begin{center}
\includegraphics[width=1.0\linewidth]{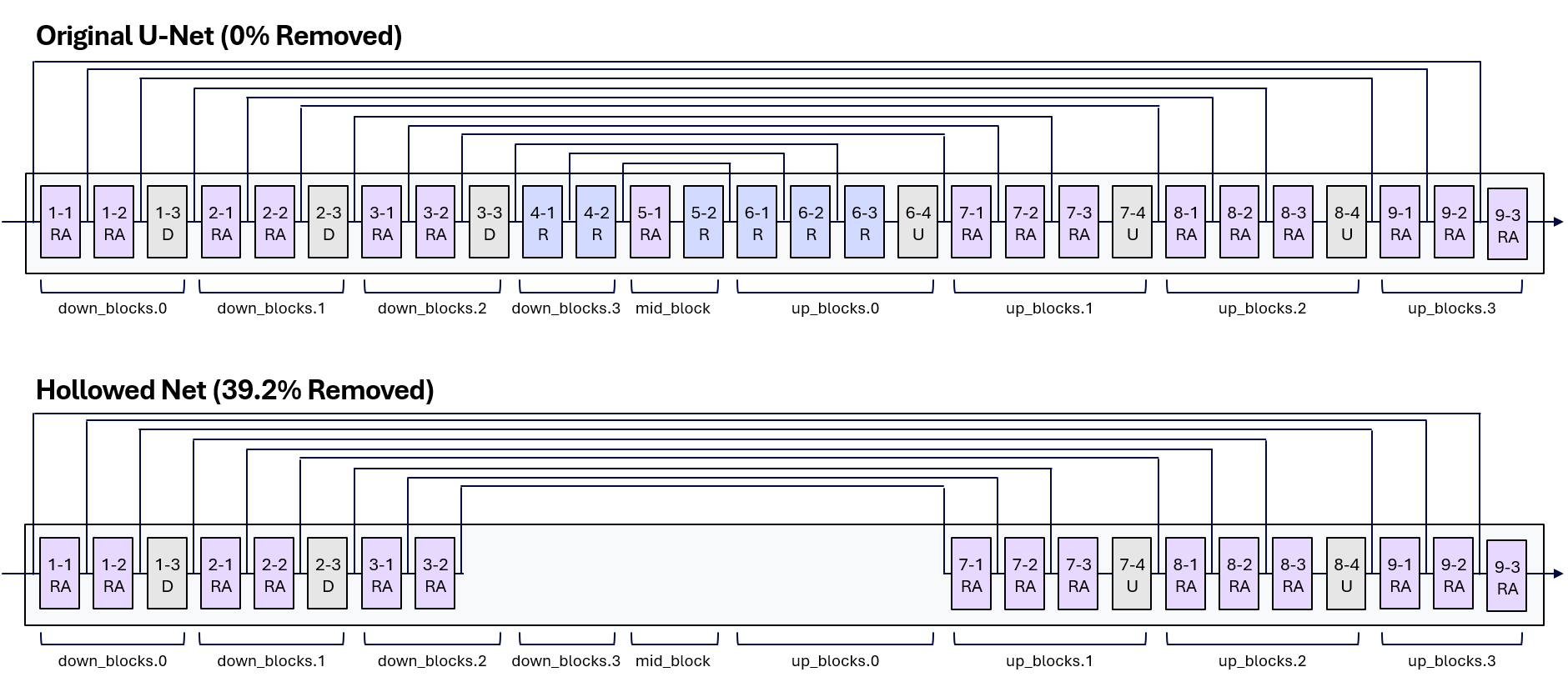}
\end{center}
   \caption{Architectural details of Hollowed Net. R represents ResBlock. RA represents a set of ResBlock and CrossAttentionBlock. D and U represent Downsample and Upsample Convs, respectively.}
\label{architecture}
\end{figure}

\section{Architectural Details of Hollowed Net}
\label{appendix_a}
In this section, we present the architectural details of Hollowed Net. We leverage the skip connections inherent in the U-Net architecture to determine which layers to be removed during fine-tuning (hollowed). For our main results, we choose the third block of the ${\rm down}\_{\rm blocks.2}$ (block 3-3), the entire ${\rm down}\_{\rm blocks.3}$ (blocks 4-1 and 4-2), the entire ${\rm mid}\_{\rm block}$ (blocks 5-1 and 5-2), and the entire ${\rm up}\_{\rm blocks.0}$ (blocks 6-1, 6-2, 6-3, and 6-4) to be hollowed, which corresponds to around 39.2\% of the U-Net's parameters, as described in Fig.~\ref{architecture}. 

Similarly, Hollowed Net with different fractions of hollowed layers can be achieved as follows:

\begin{itemize}
    \item 11.5\% removed: blocks 5-1 and 5-2 are hollowed.
    \item 20.8\% removed: blocks 4-2, 5-1, 5-2, and 6-1 are hollowed.
    \item 30.1\% removed: blocks 4-1, 4-2, 5-1, 5-2, 6-1, and 6-2 are hollowed.
    \item 56.6\% removed: blocks 3-2, 3-3, 4-1, 4-2, 5-1, 5-2, 6-1, 6-2, 6-3, 6-4, and 7-1 are hollowed.
    \item 73.3\% removed: blocks 3-1, 3-2, 3-3, 4-1, 4-2, 5-1, 5-2, 6-1, 6-2, 6-3, 6-4, 7-1, and 7-2 are hollowed.
    \item 84.3\% removed: blocks 2-3, 3-1, 3-2, 3-3, 4-1, 4-2, 5-1, 5-2, 6-1, 6-2, 6-3, 6-4, 7-1, 7-2, 7-3, and 7-4 are hollowed.
\end{itemize}

\begin{table}[h!]
\caption{Quantitative results of LoRA FT and Hollowed Net with different ranks}
\label{ranks}
\centering

\resizebox{\textwidth}{!}{
\begin{tabular}{lccccc}
\toprule
Method & \# of Parameters & Training Memory & DINO & CLIP-I & CLIP-T \\
\midrule
LoRA r=4  & 866.7M & 4.847GB & 0.564 $\pm$ 0.014 & 0.766 $\pm$ 0.006 & 0.311 $\pm$ 0.001 \\
LoRA r=16 & 869.2M & 4.883GB & 0.618 $\pm$ 0.008 & 0.788 $\pm$ 0.005 & 0.305 $\pm$ 0.001 \\
Hollow r=4  & 527.7M & 3.526GB & 0.566 $\pm$ 0.009 & 0.763 $\pm$ 0.003 & 0.311 $\pm$ 0.001 \\
Hollow r=16 & 529.9M & 3.558GB & 0.626 $\pm$ 0.009 & 0.789 $\pm$ 0.005 & 0.305 $\pm$ 0.001 \\
\bottomrule
\end{tabular}
}
\end{table}

\section{Experiments with Different Ranks}
\label{appendix_b}
In Table~\ref{ranks}, we present the results using LoRA and Hollowed Net with different ranks (4 and 16) using the DreamBooth dataset. While the default rank of 4 in the diffusers library is often used, we have found that it often oversimplifies personalization details or fails to effectively handle a range of challenging subjects and prompts. Increasing the rank from 4 to 16 improves subject fidelity. However, to achieve personalization quality comparable to full fine-tuning across all subjects and prompts, we find that the rank of 128 is necessary.

\begin{figure}[h!]
\begin{center}
\includegraphics[width=1.0\linewidth]{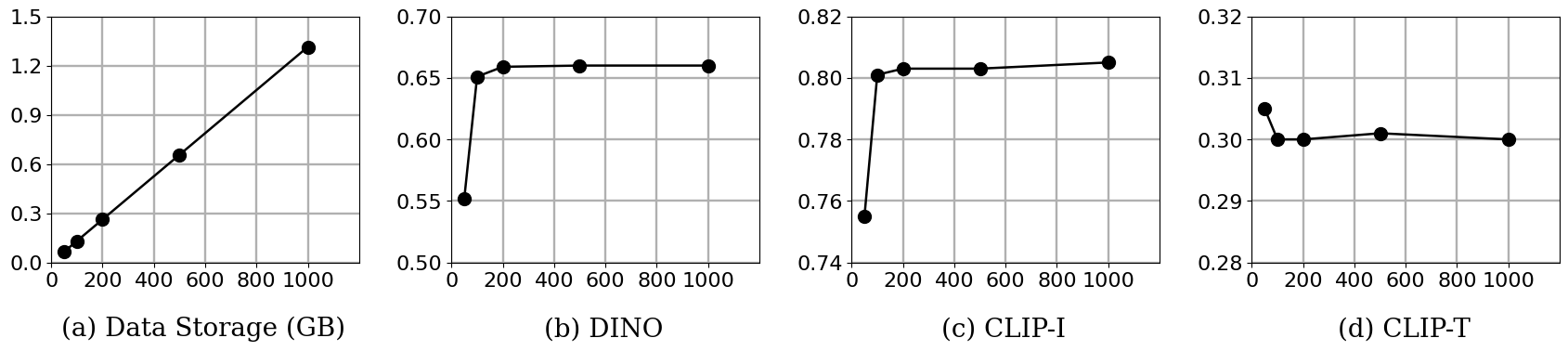}
\end{center}
   \caption{Analysis of different numbers of pre-computed samples. For all figures, the x-axis represents
the number of pre-computed samples. The y-axis corresponds to the
metric used for each figure.}
\label{collecting}
\end{figure}

\begin{table}[h!]
\centering
\caption{Resource usage analysis for different hollowed fractions and numbers of pre-computed samples}
\label{resource}
\resizebox{\textwidth}{!}{
\begin{tabular}{ccccccc}
\toprule
Hollowed Fraction & \# of Precomputed Samples & Peak GPU Usage & Date Storage & Pre-computing FLOPs & Fine-tuning FLOPs & Total FLOPs \\
\midrule
11.5\% & 200  & 4.49GB & 0.08GB & 44T  & 2081T & 2125T \\
11.5\% & 500  & 4.49GB & 0.19GB & 109T & 2081T & 2190T \\
11.5\% & 1000 & 4.49GB & 0.38GB & 218T & 2081T & 2299T \\ \midrule
39.2\% & 200  & 3.88GB & 0.26GB & 48T  & 2004T & 2052T \\
39.2\% & 500  & 3.88GB & 0.66GB & 119T & 2004T & 2123T \\
39.2\% & 1000 & 3.88GB & 1.31GB & 238T & 2004T & 2242T \\
\midrule
56.6\% & 200  & 3.16GB & 0.26GB & 56T  & 1776T & 1832T \\
56.6\% & 500  & 3.16GB & 0.66GB & 140T & 1776T & 1916T \\
56.6\% & 1000 & 3.16GB & 1.31GB & 279T & 1776T & 2055T \\
\bottomrule
\end{tabular}
}

\end{table}

\section{Further Analysis on Computational Costs}
\label{appendixC}

In Fig.~\ref{collecting}, we provide ablation studies on the impact of varying the number of samples on both quantitative and qualitative results. The findings indicate that using only 200 pre-computed samples results in minimal performance degradation compared to using 1000 pre-computed samples.

Additionally, we present a detailed analysis of computational loads and space consumption in Table~\ref{resource} for different numbers of precomputed samples and different fractions of hollowed layers, which will enable users to choose the optimal configurations of Hollowed Net according to their specific resource constraints.

\begin{figure}[h!]

  \centering
  \subfloat[DreamBooth]{\includegraphics[width=0.5\textwidth]{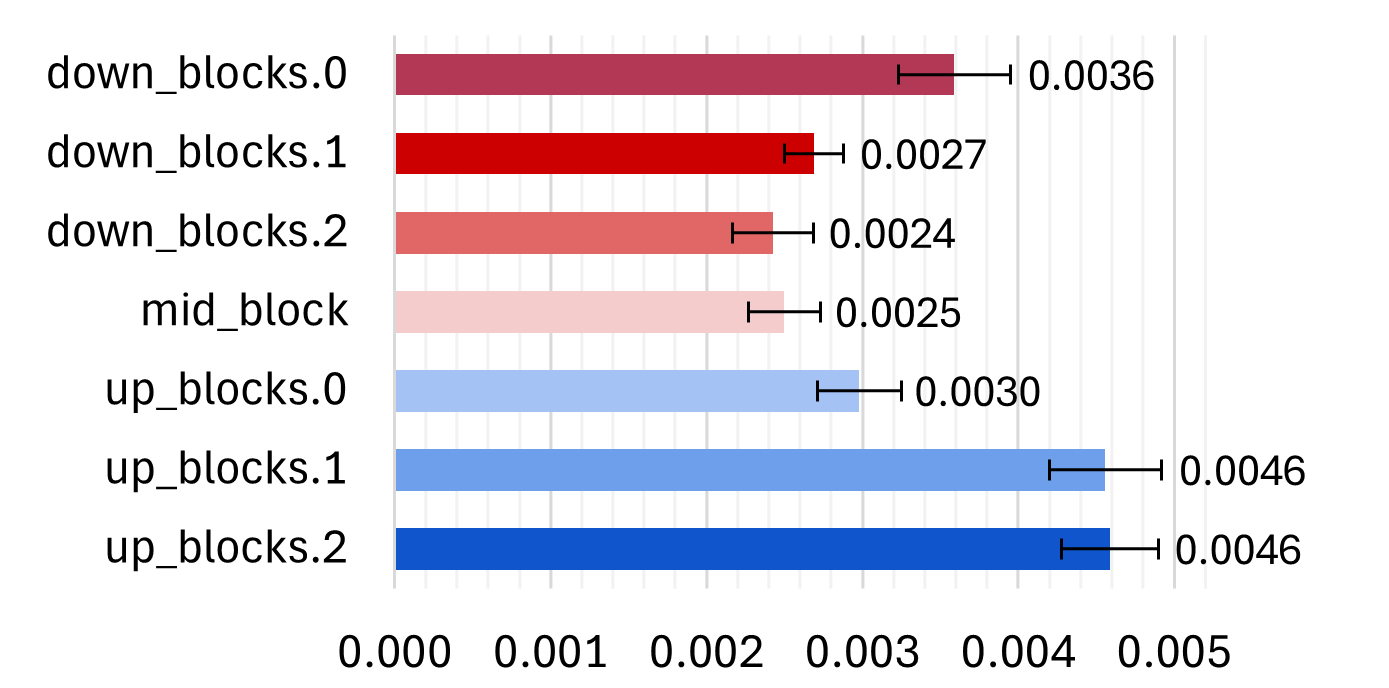}}
  \hfill
  \subfloat[CustomConcept101]{\includegraphics[width=0.5\textwidth]{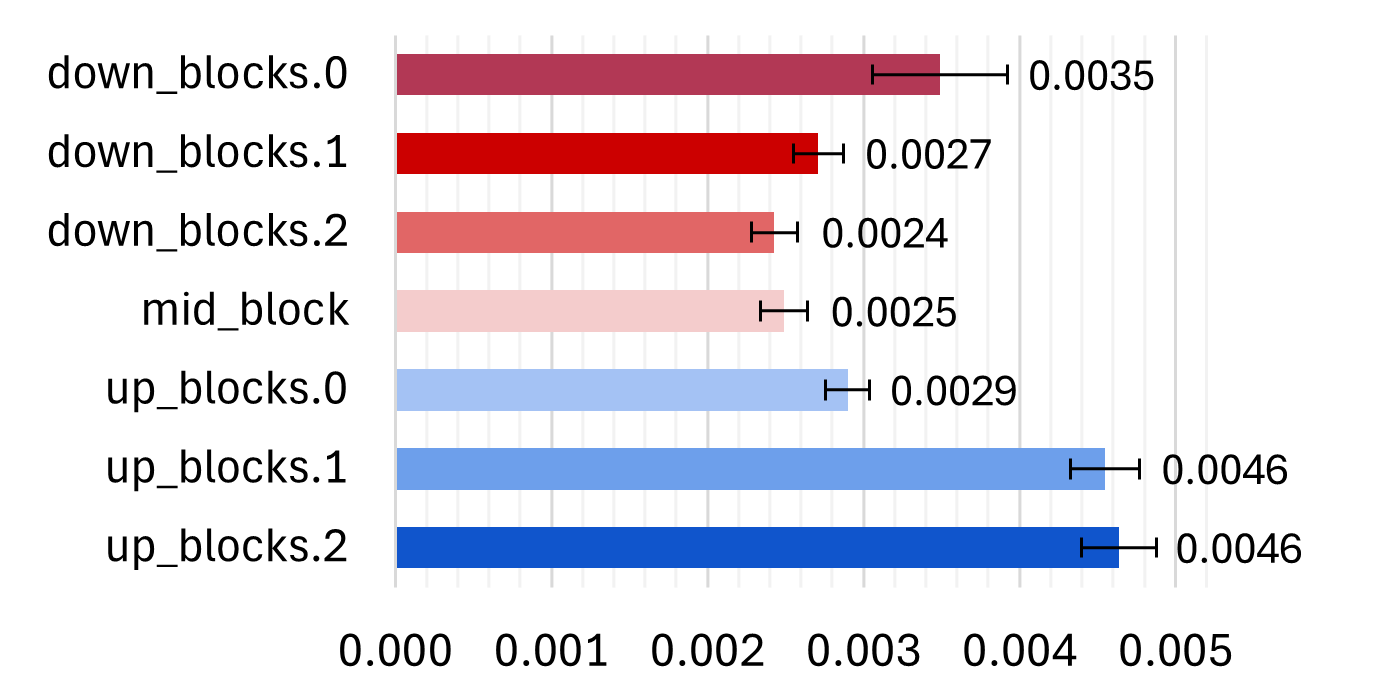}}
  \caption{Analysis of the LoRA weight change before and after personalization, per block of U-Net using SDXL~\cite{podell2023sdxl}.}
\label{sdxl_weight_change}
\end{figure}

\begin{figure}[h!]
\begin{center}
\includegraphics[width=.9\linewidth]{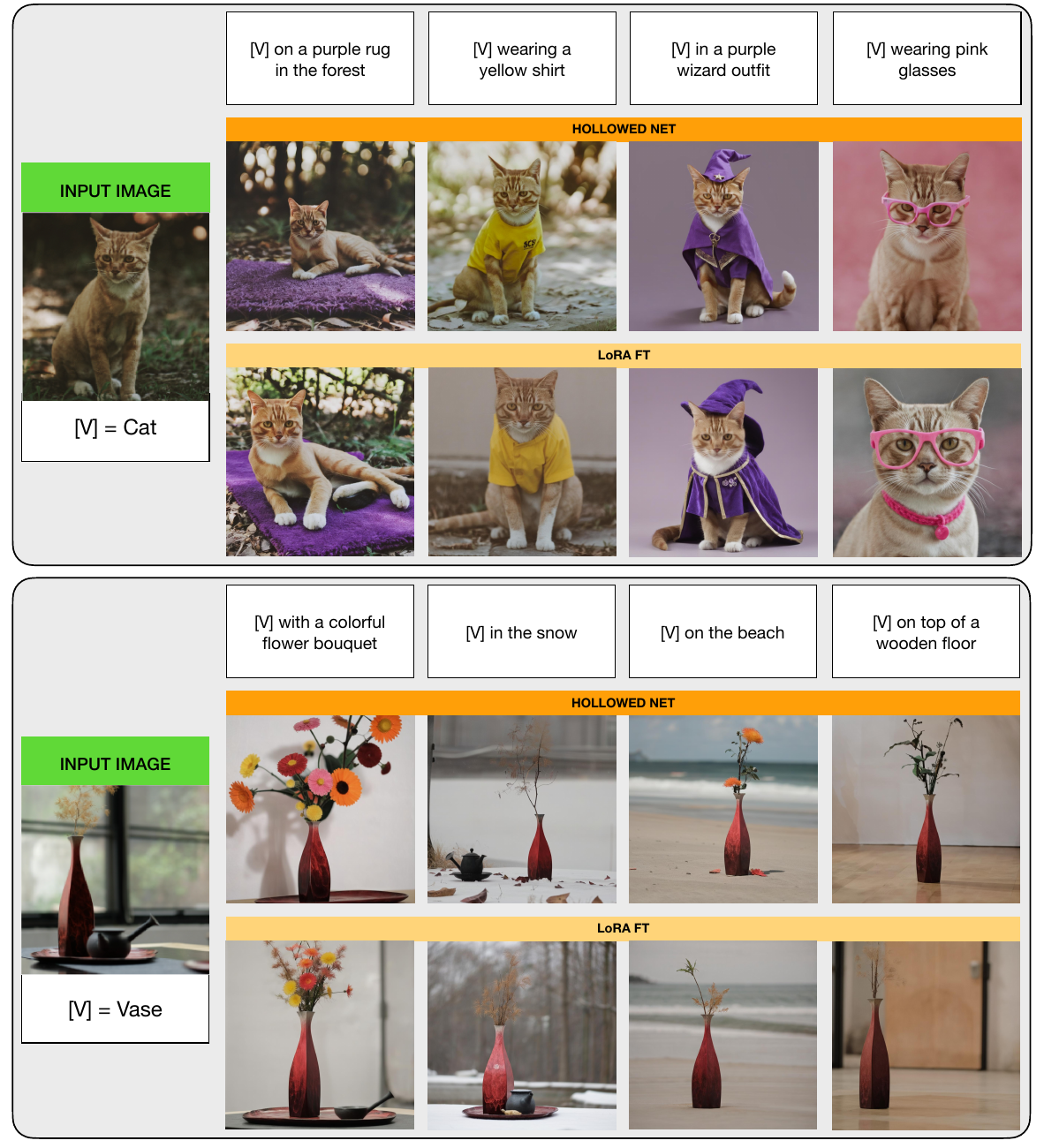}
\end{center}
   \caption{The qualitative examples of Hollowed Net and LoRA FT with the samples from the DreamBooth dataset~\cite{ruiz2023dreambooth} using SDXL~\cite{podell2023sdxl}.}
\label{sdxl_examples}
\end{figure}

\section{Experiments with SDXL}
\label{appendix_d}
To demonstrate the scalability of Hollowed Net, we present additional analysis and qualitative examples using SDXL~\cite{podell2023sdxl}. Figure~\ref{sdxl_weight_change} shows that similar patterns of weight changes are observable with SDXL, as displayed in Fig.~\ref{fig:observation}. In Fig.~\ref{sdxl_examples}, we present qualitative examples of Hollowed Net and LoRA FT with the samples from the DreamBooth dataset using SDXL. Hollowed Net is applied by removing the entire ${\rm mid}\_{\rm block}$ layers (410M parameters) of SDXL. The results show that Hollowed Net achieves high-fidelity personalization results comparable to LoRA FT.

\end{document}